\documentclass[sigconf,nonacm]{acmart}
\renewcommand\footnotetextcopyrightpermission[1]{}
\setcopyright{none}
\acmDOI{}
\acmISBN{}
\acmConference{}{}{}
\acmYear{}
\usepackage{booktabs}
\usepackage{multirow}
\usepackage{siunitx}
\usepackage[table]{xcolor}

\definecolor{m2fpblue}{RGB}{235,243,255}
\definecolor{teedgreen}{RGB}{236,248,236}
\definecolor{fusionorange}{RGB}{255,244,232}
\AtBeginDocument{%
  }

\begin{document}

\title{Sketch It Out: Exploring Label-Free Structural Cues for Multimodal Gait Recognition}

\author{Chao Zhang}
\authornote{Equal contribution.}
\email{2023020972@stu.cdut.edu.cn}
\orcid{0009-0001-4590-0358}
\affiliation{%
  \institution{Chengdu University of Technology}
  \city{Chengdu}
  \state{Sichuan}
  \country{China}
}

\author{Zhuang Zheng}
\authornotemark[1]
\email{2025020955@stu.cdut.edu.cn}
\affiliation{%
  \institution{Chengdu University of Technology}
  \city{Chengdu}
  \state{Sichuan}
  \country{China}
}
\author{Ruixin Li}
\email{lrx96@foxmail.com}
\affiliation{%
  \institution{}
  \country{}
}

\author{Zhanyong Mei}
\authornote{Corresponding author.}
\email{meizhanyong2014@cdut.edu.cn}
\affiliation{%
  \institution{Chengdu University of Technology}
  \city{Chengdu}
  \state{Sichuan}
  \country{China}
}


\begin{abstract}
Gait recognition is a non-intrusive biometric technique for security applications. Existing studies mainly rely  on silhouette, skeleton map and parsing representations.
Although silhouette captures global body shape, it is structurally sparse and lacks internal details. Skeleton maps encode different body parts but still miss fine shape cues. Parsing provides richer part-level structures, yet its effectiveness depends heavily on the quality of upstream parsing, such as label granularity and boundary precision.
These distinct characteristics suggest the need for a more unified view on gait visual representations. To this end, we revisit existing representations from the perspectives of \emph{structural edge density} and \emph{semantic richness}.
 Silhouette-based methods rely on sparse boundary structures with  single semantic foreground. Skeleton map-based methods are also structurally sparse, but encode different body parts with richer semantic information. In contrast, parsing-based methods use denser structural cues with finer semantic labels. This view reveals an underexplored direction: learning dense part-level structural information without explicit semantic labels. To fill this gap, we introduce \textbf{Sketch} as a new visual modality for gait recognition. Sketch extracts \textit{label-free}, high-frequency structural cues, such as limb articulations and self-occlusion contours, directly from RGB images using edge detectors, without explicit semantic part labels. Our experiments verify the robustness of Sketch under challenging conditions. The results further demonstrate that Sketch and Parsing effectively complement each other: Sketch compensates for structural boundaries missing in Parsing, while Parsing mitigates noising texture and other irrelevant patterns over-extracted by edge-based Sketch. To better exploit the complementary characteristics of these two modalities, we propose \textbf{SketchGait}, a multi-modal framework with two independent streams for modality-specific learning and a lightweight early-fusion branch. Furthermore, by incorporating the \textbf{Silhouette} modality, we extend it to \textbf{SketchGait++}, which further improves the overall performance.
Extensive experiments on SUSTech1K and CCPG demonstrate the effectiveness of the proposed modality and framework.
\end{abstract}

\maketitle

\begin{figure}[h]
  \centering
  \includegraphics[width=\linewidth]{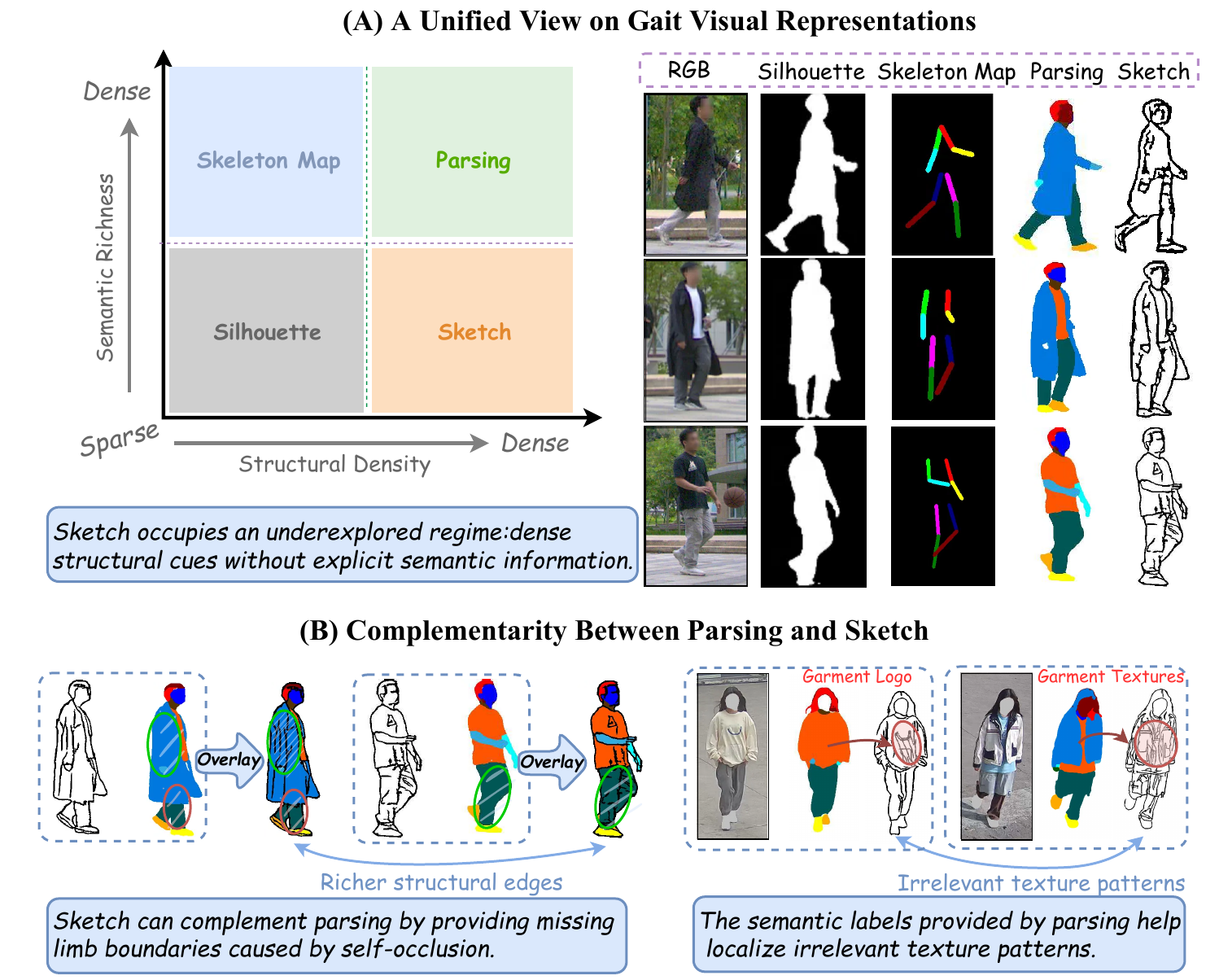}
\caption{Representation analysis and complementarity of gait visual modalities.
(A) A unified view of gait representations based on structural edge density and semantic richness, covering silhouette, skeleton map, parsing, and sketch. Notably, sketch occupies an underexplored regime with dense structural cues and no explicit semantic labels.
(B) Complementarity between parsing and sketch: sketch recovers fine limb boundaries that are not well captured by the parsing modality under self-occlusion, while parsing labels help localize irrelevant clothing texture patterns for sketch regularization.}

  \Description{A woman and a girl in white dresses sit in an open car.}
\label{fig:fig_1}
\end{figure}
\section{Introduction}
Gait recognition aims to identify individuals from their walking patterns at a distance, making it a non-intrusive biometric technique for real-world security applications \cite{Bouchrika2017Survey}. A core problem in gait recognition is how to build an effective visual representation. Among existing choices, \textbf{\textit{silhouette-based}} representations have been the dominant modality for a long time. By representing a person as a binary mask, silhouettes can effectively suppress background interference. However, they are inherently sparse and preserve only the outer contour of the human body, while discarding internal structural details. As a result, fine-grained cues, such as the relative motion of body parts, are often lost, which limits the ability of the model to capture subtle but discriminative gait patterns.

Several works convert skeleton sequences into \textit{\textbf{skeleton maps}} \cite{fan2024skeletongait,xu2025psgait,fu2024cut}. A representative example is the \emph{Parsing Skeleton} modality, which renders skeletons as 2D maps and assigns different semantics to different skeletal segments for richer representation. However, due to the sparsity of skeleton points, it mainly captures internal body structure while lacking informative outer contours and structural details of different body parts, which limits its further development.

Recent studies have introduced \textbf{\textit{parsing-based}} representations for gait recognition \cite{zheng2023parsing}. By decomposing the human body into semantic parts, parsing provides richer internal structures and part-level semantic cues than silhouettes. With these fine-grained representations, parsing-based methods have achieved strong performance on in-the-wild gait datasets \cite{zheng2023parsing,zheng2022gait}. However, current parsing-based methods in gait recognition still have notable limitations. Even under the same backbone, parsing-based models do not always outperform silhouette-based ones, and may even perform worse in some cases \cite{jin2025exploring}. This is largely because their performance depends heavily on the quality of the upstream parser and the granularity of part labels. Segmentation errors and coarse semantic partitions can reduce the reliability of both body boundaries and internal structures.
These issues can be alleviated by high-quality parsing maps. Stronger parsing models  provide more reliable structural cues and often lead to more stable performance \cite{yang2024deep}. Nevertheless, parsing still relies on predefined semantic labels and mainly represents the body at a middle-level structural granularity, which may cause local inconsistencies under inaccurate predictions and miss finer structural details beyond semantic part partitions.

The different characteristics of these visual modalities motivate us to reconsider them from a unified perspective. From this view, as illustrated in Fig.~\ref{fig:fig_1}(A), we revisit existing gait representation modalities along two dimensions: \textbf{structural edge density}, ranging from sparse to dense, and \textbf{semantic richness}, ranging from limited to rich. Silhouette-based representations describe the outer body shape as a binary mask with a single foreground label, and thus provide sparse structural cues with weak semantic information. Skeleton map-based representations preserve only the internal body structure and lack sufficient shape information, but different bones still encode rich semantics for different body parts. In contrast, parsing-based representations provide denser part-level structures with stronger semantic information through multiple predefined semantic labels. These reveal an underexplored direction: preserving dense part-level structural information without explicit semantic information. 

To fill this gap, we introduce \textbf{Sketch} as a new visual modality for gait recognition, as shown in Fig.~\ref{fig:fig_1}(A). Here, Sketch denotes an edge-based structural representation that preserves sketch-like high-frequency human cues without explicit semantic part labels. It is designed to capture fine-grained structural details, such as internal boundaries and self-occlusion contours. Compared with parsing, Sketch provides denser structural boundaries while avoiding the  label-supervised paradigm required by parsing, making it a completely \emph{\textbf{label-free}} representation. In practice, \textit{Sketch} is mainly constructed by applying edge detectors to RGB images, including Sobel \cite{sobel19683x3}, Canny \cite{canny2009computational}, PiDiNet \cite{su2021pixel}, and TEED \cite{soria2023tiny}. It can also be extended to other forms, where structured cues such as parsing maps are converted into edge-based representations and treated as coarse-grained \textit{Sketch}.

Sketch representations generated by edge-based detectors have been explored in other vision tasks, such as sketch-based image retrieval \cite{9093290} and medical image segmentation \cite{du2022net}, but their use in gait recognition remains limited.  Compared with parsing-based representations, Sketch has two potential advantages. First, Sketch is \emph{label-free}. It does not rely on predefined semantic labels and is thus less constrained by \emph{label-guided} priors. This can be helpful on imbalanced datasets \cite{shen2023lidargait}, where strong semantic priors may introduce spurious correlations with identity. Second, Sketch preserves high-frequency edges directly from RGB images. Since it does not require semantic pixel assignment, it is less affected by the self-occlusion ambiguity in parsing, where overlapping body parts may be merged into the same label and lose useful motion cues.

However, Sketch also has its limitations. Current deep learning-based edge detectors \cite{su2021pixel,soria2023tiny} are not designed for gait recognition. As a result, they may fail to distinguish human structural edges from irrelevant appearance details. For example, as shown in Fig.~\ref{fig:fig_1}(B), they may over-detect clothing logos and texture patterns. These irrelevant edges can interfere with feature learning and may introduce spurious correlations with irrelevant covariates, which can hurt generalization.

To address these limitations, we further explore the complementary roles of Parsing and Sketch. While Sketch preserves richer structural details, Parsing provides more stable semantic structure, which may help regularize Sketch when irrelevant edges are over-detected, as illustrated in Fig. 1(B). These observations suggest that the two modalities offer complementary structural and semantic information. Based on this idea, we propose \textbf{SketchGait}, a multi-modal gait recognition framework that jointly models Parsing and Sketch. It uses two independent streams and a lightweight early fusion branch to capture complementary cues. We further extend it to \textbf{SketchGait++} by introducing the silhouette modality. This design enables effective interaction among modalities and improves gait recognition performance.

\textbf{Our main contributions are summarized as follows:}
\begin{itemize}
\item \textbf{Representation-level analysis.}
We revisit common gait representations from two perspectives: \emph{structural edge density} and \emph{semantic richness}. This view reveals an underexplored direction: dense structural cues without explicit semantic information. Therefore, we introduce \emph{Sketch} as a new \textit{label-free} structural modality for gait recognition.

\item \textbf{SketchGait and SketchGait++ framework.}
Based on the complementary properties of \textit{Parsing} and \textit{Sketch}, we propose SketchGait, a dual-stream framework with modality-specific backbones and a lightweight early fusion branch. By further introducing the silhouette modality, we extend it to SketchGait++.

  \item \textbf{Extensive experiments.}
  We conduct extensive experiments on SUSTech1K and CCPG. The results verify the effectiveness of the Sketch modality and demonstrate the advantage of SketchGait over strong baselines.
\end{itemize}

\section{Related Work}
\subsection{Visual Representations for Gait Recognition}
Gait recognition largely depends on the choice of visual representation, which affects how well motion patterns and structural cues can be captured. Existing studies have explored different visual modalities with different trade-offs between robustness and information richness.

\textbf{Silhouette-based Representations.}
Binary silhouettes have long been the dominant representation in gait recognition due to their simplicity and robustness to background clutter. Early methods aggregated silhouettes into static templates such as the Gait Energy Image (GEI) \cite{han2005individual}, while recent deep models \cite{chao2019gaitset,fan2020gaitpart,lin2022gaitgl} process silhouette sequences or sets to learn spatiotemporal features. Despite their effectiveness, silhouettes preserve only coarse body shape and discard internal structural cues, which limits their ability to model fine-grained gait dynamics in complex scenarios.

\textbf{Skeleton-based Representations.}
Skeleton-based methods \cite{deng2018human,teepe2021gaitgraph,catruna2024gaitpt} represent gait using 2D human keypoints extracted by pose estimation models. They explicitly model body kinematics and are often robust to view and clothing variations. However, skeleton representations are inherently sparse and abstract, lacking detailed body shape and fine-grained structural information. Even with recent map-based formulations \cite{fan2024skeletongait,xu2025psgait,zheng2024takes}, they remain limited in modeling body contours and non-rigid structural variations of limbs and body parts, which are useful for robust gait recognition.

\textbf{Parsing-based Representations.}
Human parsing has recently been introduced as a semantically enriched gait representation by segmenting the human body into pixel-level semantic parts. By providing label-guided semanitc priors, parsing-based methods \cite{zheng2023parsing,jin2025exploring} enable part-level feature learning and have achieved strong performance on in-the-wild benchmarks \cite{zheng2022gait}. However, their effectiveness still depends on the quality of the upstream parser and the granularity of semantic part labels. More importantly, parsing mainly represents the body through predefined semantic partitions, which may limit its ability to capture finer structural details beyond semantic parts.

\subsection{Edge Detection and Sketch Extraction}
Edge detection is a fundamental low-level vision task that aims to extract object boundaries by identifying intensity discontinuities. Traditional operators such as Canny \cite{canny2009computational} and Sobel \cite{sobel19683x3} rely on gradient and are sensitive to noise and texture, often producing cluttered and irrelevant edges in complex scenes, which limits their applicability for gait representation.

\textbf{Deep Learning-based Edge Detection.}
With the advent of convolutional neural networks, edge detection has shifted from gradient-based filtering to data-driven feature learning. Early works such as Holistically-Nested Edge Detection (HED) \cite{xie2015holistically} exploit multi-scale supervision to extract boundaries, while subsequent methods including RCF \cite{liu2017richer} and BDCN \cite{he2019bi} further improve localization accuracy through deep feature aggregation. More recent lightweight models, such as PiDiNet \cite{su2021pixel} and TEED \cite{soria2023tiny}, focus on efficient boundary extraction and achieve a favorable trade-off between computational efficiency and edge quality.

\textbf{Sketch as a Gait Modality.}
In this work, we construct the Sketch modality for gait recognition using edge detectors, including deep models such as TEED. Compared with binary silhouettes and \textit{label-guided} parsing maps, Sketch provides denser and more fine-grained structural cues without explicit semantic labels, such as limb contours and self-occlusion boundaries. This \textit{label-free} representation helps preserve raw structural motion patterns beyond predefined semantic partitions, making Sketch a useful complementary modality for gait recognition under complex conditions.

\subsection{Multi-modal Fusion in Gait Recognition}
Multi-modal fusion has been widely used in gait recognition to compensate for the limitations of individual visual modalities by integrating complementary information.

From a methodological perspective, existing methods can be roughly divided into \textbf{\emph{heterogeneous}} and \textbf{\emph{homogeneous}} fusion. Heterogeneous fusion focuses on combining fundamentally different modalities, such as skeleton sequences with silhouettes or parsing maps. Representative methods, such as BiFusion \cite{peng2024learning} and TriGait \cite{sun2023trigait}, usually adopt late fusion or attention-based interactions to align heterogeneous features across modalities.

In contrast, homogeneous fusion explores the joint modeling of closely related visual modalities within the same domain. SkeletonGait++ \cite{fan2024skeletongait} improves robustness through early integration of 2D skeleton maps and silhouettes. XGait \cite{zheng2024takes} further combines human parsing and silhouettes to perform spatial semantic alignment across different granularity levels. More recently, MultiGait++ \cite{jin2025exploring} explicitly models shared and modality-specific features to capture complementary information among multiple visual modalities.

\begin{figure*}[h]

  \centering
  \includegraphics[width=\linewidth]{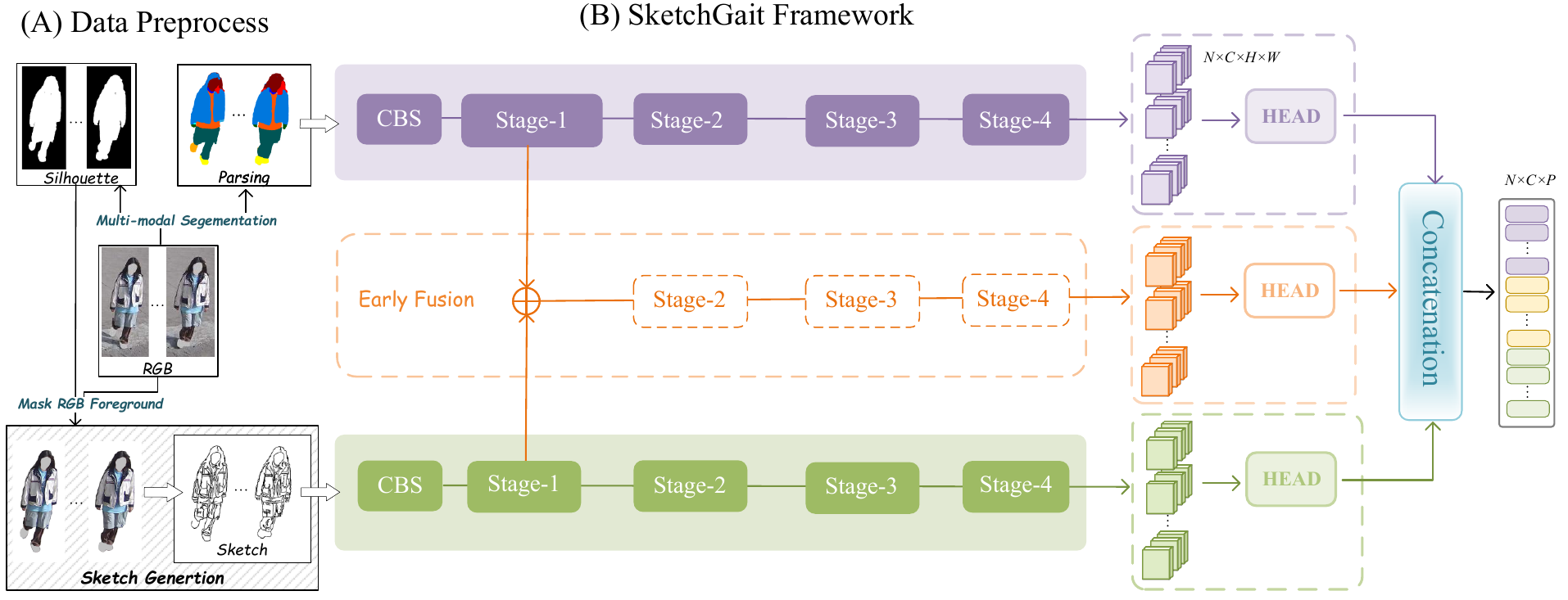}
  
\caption{
Overall pipeline. 
(A) Data Preprocess. 
(B) The SketchGait framework with two modality-specific input branches and one fusion branch. 
CBS denotes a Conv $3\times3$ layer followed by BatchNorm and ReLU. 
Stage 1--4 correspond to different stages of the DeepGaitV2 backbone. 
The HEAD module consists of temporal max pooling, horizontal pyramid pooling (HPP), 
and separate fully connected (FC) layers for feature projection.
}
\label{fig:framework}
\end{figure*}

\section{Method}
\subsection{Sketch Modality Construction}
The Sketch modality is designed to capture dense, \textit{label-free} structural cues with reduced background interference. However, directly applying edge detectors, such as Sobel or TEED, to raw RGB frames often introduces many background edges, which may distract gait feature learning. As shown in Fig.~\ref{fig:framework}(A), we therefore directly remove the background using the foreground mask before Sketch construction. Let $I \in \mathbb{R}^{H \times W \times 3}$ denote an input RGB frame, and let $M \in \{0,1\}^{H \times W}$ denote a binary foreground mask indicating the human region. The mask $M$ can be obtained either from the silhouette modality or from the union of human semantic regions in the parsing modality. We first extract the foreground RGB image by
\begin{equation}
I_{\text{fg}} = I \odot M,
\end{equation}
where $\odot$ denotes element-wise multiplication. We then apply a pre-trained edge detector $\mathcal{F}_{\text{edge}}$ to the masked foreground to obtain the Sketch representation:
\begin{equation}
S = \mathcal{F}_{\text{edge}}(I_{\text{fg}}).
\end{equation}

By suppressing most background content before edge extraction, Sketch retains human contours and internal structures, including limb boundaries and self-occlusion edges. This produces a dense structural representation that complements parsing-based modalities. Although more advanced preprocessing may lead to higher-quality Sketch representations, the goal of this work is not to optimize preprocessing, but to investigate the value of Sketch for gait recognition through a comprehensive empirical study.

\subsection{SketchGait Framework}
Based on the constructed Sketch modality, we propose \textbf{SketchGait}, a multi-modal gait recognition framework that jointly models Parsing and Sketch, as illustrated in Fig.~\ref{fig:framework}(B). The framework is built on a simple  principle: allowing shallow cross-modal interaction while preserving modality-specific learning at deeper stages.

\textbf{Design Motivation.}
SketchGait is motivated by two observations.  First, the two modalities correspond to different semantic forms: Parsing is \textit{label-guided}, whereas Sketch is \textit{label-free}. This difference imposes different semantic constraints on representation learning, and may lead the two modalities to focus on different structural patterns. Second, Sketch preserves fine-grained edges and high-frequency contours that are often less explicit in Parsing, while Parsing provides more stable part-level semantic structure.

\textbf{Framework Design.}
Based on these observations, SketchGait adopts a dual-stream architecture with weight-independent backbones to preserve modality-specific learning. To model shared structural information, we further introduce a lightweight early-stage fusion branch. Empirically, early interaction shows a slight advantage over late fusion in our setting. Therefore, SketchGait performs fusion at shallow stages and keeps the two streams independent at deeper stages.


\textbf{SketchGait Formulation.}
Let $X_{\text{ske}}$ and $X_{\text{par}}$ denote the Sketch and Parsing inputs. The two modality-specific streams first extract shallow features at Stage-1:
\begin{equation}
F_{\text{ske}}^{(1)} = \mathcal{B}_{\text{ske}}^{(1)}(X_{\text{ske}}), \quad
F_{\text{par}}^{(1)} = \mathcal{B}_{\text{par}}^{(1)}(X_{\text{par}}),
\end{equation}
where $\mathcal{B}_{(\cdot)}^{(1)}$ denotes the Stage-1 backbone block. We then build a shallow fusion branch by
\begin{equation}
F_{\text{fus}}^{(1)} = F_{\text{ske}}^{(1)} + F_{\text{par}}^{(1)}.
\end{equation}

The three branches are processed independently in the remaining backbone stages, producing feature maps $F_{\text{ske}}$, $F_{\text{par}}$, and $F_{\text{fus}}$. For each branch, temporal max pooling, Horizontal Pyramid Pooling (HPP), and branch-specific fully connected layers are used to obtain the embedding
\begin{equation}
\mathbf{e}_{b} = \mathrm{FC}_{b}\big(\mathcal{H}(\mathcal{P}(F_{b}))\big), \quad
b \in \{\text{ske}, \text{par}, \text{fus}\},
\end{equation}
where $\mathcal{P}(\cdot)$ and $\mathcal{H}(\cdot)$ denote temporal max pooling and HPP, respectively.

The final embedding is obtained by
\begin{equation}
\mathbf{e}_{\text{final}} = \mathrm{Concat}(\mathbf{e}_{\text{ske}}, \mathbf{e}_{\text{par}}, \mathbf{e}_{\text{fus}}).
\end{equation}

During training, each branch embedding is fed into an independent BNNeck layer to produce logits $\mathbf{z}_{\text{ske}}$, $\mathbf{z}_{\text{par}}$, and $\mathbf{z}_{\text{fus}}$. The final logit is
\begin{equation}
\mathbf{z}_{\text{final}} = \mathrm{Concat}(\mathbf{z}_{\text{ske}}, \mathbf{z}_{\text{par}}, \mathbf{z}_{\text{fus}}).
\end{equation}

This design preserves modality-specific learning at deeper stages while modeling shallow structural complementarity between Parsing and Sketch modalities.
\subsection{SketchGait++ Framework}
To further improve overall performance, we incorporate the silhouette modality into the SketchGait framework. Silhouette provides stable global shape cues and is naturally complementary to \textit{Sketch} and \textit{Parsing}. Based on this, we build \textbf{SketchGait++} by concatenating \textit{Silhouette} with \textit{Sketch} at the Sketch input along the channel dimension.
\subsection{Loss Function}

Following common practice in gait recognition, SketchGait is optimized with a joint objective that combines metric learning and identity classification. Specifically, we employ the Triplet Loss to enforce inter-class separability in the embedding space, together with the Cross-Entropy Loss to provide direct identity supervision. The overall training objective is formulated as
\begin{equation}
\mathcal{L} = \mathcal{L}_{\text{tri}} + \mathcal{L}_{\text{ce}},
\end{equation}
where $\mathcal{L}_{\text{tri}}$ is the batch-hard triplet loss that encourages samples from the same identity to be closer than those from different identities, and $\mathcal{L}_{\text{ce}}$ denotes the cross-entropy loss computed on the corresponding classification logits.

\begin{table*}[t]
  \centering
  \caption{Results under different input modalities on SUSTech1K.}
  \label{tab:modalities_results}

  \renewcommand{\arraystretch}{1.12}

  \begin{tabular}{ll
                  S[table-format=2.1] S[table-format=2.1] S[table-format=2.1] S[table-format=2.1]
                  S[table-format=2.1] S[table-format=2.1] S[table-format=2.1] S[table-format=2.1]}
    \toprule
    \multirow{3}{*}{Modality} & \multirow{3}{*}{Source} &
    \multicolumn{8}{c}{Backbone Method} \\
    \cmidrule(lr){3-10}
    & &
    \multicolumn{4}{c}{GaitBase \cite{fan2023opengait}} &
    \multicolumn{4}{c}{DeepGaitV2 \cite{fan2023exploring}}  \\
    \cmidrule(lr){3-6}\cmidrule(lr){7-10}
    & &
    {NM} & {CL} & {NT} & {OA@R1} &
    {NM} & {CL} & {NT} & {OA@R1} \\
    \midrule

    \multirow{6}{*}{Silhouette}
    & \textit{Original Silh.} & \multicolumn{8}{c}{} \\
    & Official Released
      & 81.5 & 49.6 & 26.0 & 76.1
      & 86.5 & 49.0 & 28.0 & 80.9 \\
    \addlinespace[2pt]
    & & \multicolumn{8}{c}{\cellcolor{m2fpblue}\textit{Pars.-to-Silh.}} \\
    & GaitParsing \cite{wang2023gaitparsing}
      & {\cellcolor{m2fpblue}74.1} & {\cellcolor{m2fpblue}45.5} & {\cellcolor{m2fpblue}56.2} & {\cellcolor{m2fpblue}68.5}
      & {\cellcolor{m2fpblue}79.0} & {\cellcolor{m2fpblue}47.8} & {\cellcolor{m2fpblue}61.1} & {\cellcolor{m2fpblue}74.8} \\
    & CDGNet \cite{liu2022cdgnet}
      & {\cellcolor{m2fpblue} 77.5} & {\cellcolor{m2fpblue}31.3} & {\cellcolor{m2fpblue}43.5} & {\cellcolor{m2fpblue}69.2}
      
      & {\cellcolor{m2fpblue}82.2} & {\cellcolor{m2fpblue}32.9} & {\cellcolor{m2fpblue} 49.8} & {\cellcolor{m2fpblue}76.4} \\
    & M2FP \cite{yang2024deep}
      & {\cellcolor{m2fpblue}81.7} & {\cellcolor{m2fpblue}44.3} & {\cellcolor{m2fpblue}42.8} & {\cellcolor{m2fpblue}76.9}
      & {\cellcolor{m2fpblue}86.3} & {\cellcolor{m2fpblue}47.0} & {\cellcolor{m2fpblue}46.2} & {\cellcolor{m2fpblue}82.5} \\
    \midrule

 \multirow{2}{*}{Skeleton Map}
& Parsing Skeleton \cite{xu2025psgait}
  & 86.1 & 55.0 & 47.1 & 83.4
  & \multicolumn{1}{c}{--} & \multicolumn{1}{c}{--} & \multicolumn{1}{c}{--} & \multicolumn{1}{c}{--} \\
& SkeletonGait skeleton map\cite{fan2024skeletongait}
  & \multicolumn{1}{c}{--} & \multicolumn{1}{c}{--} & \multicolumn{1}{c}{--} & \multicolumn{1}{c}{--}
  & 55.0 & 24.7 & 43.9 & 50.1 \\
  \midrule

    \multirow{3}{*}{Parsing}
    & GaitParsing
      & 79.8 & 41.7 & 60.1 & 72.3
      & 83.3 & 41.0 & 62.5 & 78.2 \\
    & CDGNet
      & 81.6 & 23.8 & 50.0 & 73.8
      & 86.6 & 25.0 & 53.8 & 79.3 \\
    & M2FP
      & 90.8 & 41.2 & 46.0 & 84.6
      & 93.1 & 41.0 & 47.5 & 87.5 \\
    \midrule

    \multirow{9}{*}{Sketch}
    & & \multicolumn{8}{c}{\cellcolor{m2fpblue}\textit{Pars.-to-Edge}} \\
    & GaitParsing
      & {\cellcolor{m2fpblue}77.4} & {\cellcolor{m2fpblue}41.4} & {\cellcolor{m2fpblue}57.0} & {\cellcolor{m2fpblue}69.9}
      & {\cellcolor{m2fpblue}81.7} & {\cellcolor{m2fpblue}40.3} & {\cellcolor{m2fpblue}60.6} & {\cellcolor{m2fpblue}76.0} \\
    & CDGNet
      & {\cellcolor{m2fpblue}78.7} & {\cellcolor{m2fpblue}23.9} & {\cellcolor{m2fpblue}43.5} & {\cellcolor{m2fpblue}69.3}
      & {\cellcolor{m2fpblue}82.0} & {\cellcolor{m2fpblue}24.8} & {\cellcolor{m2fpblue}48.2} & {\cellcolor{m2fpblue}75.7} \\
    & M2FP
      & {\cellcolor{m2fpblue}88.7} & {\cellcolor{m2fpblue}41.4} & {\cellcolor{m2fpblue}46.5} & {\cellcolor{m2fpblue}82.6}
      & {\cellcolor{m2fpblue}90.3} & {\cellcolor{m2fpblue}43.0} & {\cellcolor{m2fpblue}49.8} & {\cellcolor{m2fpblue}86.2} \\
    \addlinespace[2pt]
    & & \multicolumn{8}{c}{\cellcolor{teedgreen}\textit{Edge-based}} \\
    & Canny \cite{canny2009computational}
      & {\cellcolor{teedgreen}75.4} & {\cellcolor{teedgreen}36.2} & {\cellcolor{teedgreen}38.0} & {\cellcolor{teedgreen}68.2}
      & {\cellcolor{teedgreen} 81.4} & {\cellcolor{teedgreen}38.4} & {\cellcolor{teedgreen}}38.4& {\cellcolor{teedgreen}74.3} \\
    & Sobel \cite{sobel19683x3}
      & {\cellcolor{teedgreen}91.7} & {\cellcolor{teedgreen}53.2} & {\cellcolor{teedgreen}49.3} & {\cellcolor{teedgreen}85.6}
      & {\cellcolor{teedgreen}91.8} & {\cellcolor{teedgreen}52.6} & {\cellcolor{teedgreen}49.0} & {\cellcolor{teedgreen}87.1} \\
    & PidiNet \cite{su2021pixel}
      & {\cellcolor{teedgreen}84.3} & {\cellcolor{teedgreen}50.3} & {\cellcolor{teedgreen}38.2} & {\cellcolor{teedgreen}75.7}
      & {\cellcolor{teedgreen}82.5} & {\cellcolor{teedgreen}53.3} & {\cellcolor{teedgreen}37.2} & {\cellcolor{teedgreen}77.8} \\
    & TEED \cite{yang2024deep}
      & {\cellcolor{teedgreen}\textbf{93.3}} & {\cellcolor{teedgreen}\textbf{57.9}} & {\cellcolor{teedgreen}\textbf{49.0}} & {\cellcolor{teedgreen}\textbf{87.6}}
      & {\cellcolor{teedgreen}\textbf{94.5}} & {\cellcolor{teedgreen}\textbf{56.0}} & {\cellcolor{teedgreen}\textbf{51.5}} & {\cellcolor{teedgreen}\textbf{89.6}} \\
    \bottomrule
  \end{tabular}
\end{table*}

\section{Experiments}
\subsection{Datasets}
This paper investigates and analyzes the \emph{edge-based Sketch} modality for gait recognition. To the best of our knowledge, existing gait benchmarks have not evaluated Sketch-style representations, leaving this direction unexplored. Since our edge-based Sketch detector requires \textbf{RGB} inputs and benefits from higher image resolution for more accurate edge extraction, we restrict our study to large-scale public datasets with RGB data for fair comparison and validation.

\textbf{SUSTech1K} \cite{shen2023lidargait} is a recently proposed large-scale \emph{outdoor} dataset with diverse and practical covariates. It contains multiple gait conditions, including Normal (NM), Bags (BG), Clothing (CL), Carrying (CR), Umbrella (UM), Uniform (UN), Occlusion (OC), and Night-time (NT). SUSTech1K includes \textbf{1,050} subjects in total, where \textbf{250} subjects are used for training and the remaining \textbf{800} subjects are used for testing.

As a second benchmark, we use \textbf{CCPG} \cite{li2023depth}, which consists of both \emph{indoor} and \emph{outdoor} scenes and provides rich clothing variations for each subject. Specifically, the dataset covers full clothing changes (CL-FULL), upper-body changes (CL-UP), lower-body changes (CL-DOWN), as well as backpack-related variations (CL-BG). CCPG contains \textbf{200} subjects, with the first \textbf{100} used for training and the remaining \textbf{100} used for testing.

\begin{table*}[t]
  \centering
  \caption{Performance comparison on SUSTech1K, where Parsing and Sketch are generated by M2FP and TEED, respectively.}
  \label{tab:sustech1k_comparsion}
  \setlength{\tabcolsep}{4pt}
  \renewcommand{\arraystretch}{1.1}
  \begin{tabular}{llcccccccccc}
    \toprule
    \textbf{Input} & \textbf{Method} &
    \textbf{NM} & \textbf{BG} & \textbf{CL} & \textbf{CR} &
    \textbf{UM} & \textbf{UN} & \textbf{OC} & \textbf{NT} &
    \textbf{OA@\textit{R1}} & \textbf{OA@\textit{R5}} \\
    \midrule

    \multirow{5}{*}{Silhoueete}
      & GaitSet \cite{chao2019gaitset}         & 69.1 & 68.2 &  37.4 & 65.0   & 63.1    & 61.0 & 67.2 & 23.0 &65.0 & 84.8    \\
      & GaitPart \cite{fan2020gaitpart}         & 62.2  & 62.8 &  33.1 & 59.5  & 57.2 & 54.8 & 57.2 & 21.7 &59.2 & 80.8    \\
      & GaitGL \cite{lin2022gaitgl}         & 67.1 & 66.2 &  35.9 & 63.3  & 61.6 & 58.1 & 66.6 & 17.9 &63.1 & 82.8    \\
      & GaitBase \cite{fan2023opengait}         & 81.5 & 77.5 &  49.6 & 75.8  & 75.5 & 76.7 & 81.4 & 25.9 &76.1 & 89.4    \\
      & DeepGaitV2 \cite{fan2023exploring}         & 86.5 & 82.8 &  49.2 & 80.4   & 83.3 & 81.9 & 86.0 & 28.0 &80.9 & 91.9    \\
    \midrule
    \multirow{1}{*}{Parsing}
      & DeepGaitV2       & 93.1 & 87.9 &  40.9 & 88.8  & 87.7 & 88.7 & 93.5 & 47.5 &87.5 & 95.1    \\
    \midrule
    \multirow{1}{*}{Sketch}
      & DeepGaitV2         &  94.5 & 90.1 &  56.4 & 90.5  & 88.1 & 89.3 & 97.3 & 51.5 &89.6 & 96.3    \\
    \midrule
    \multirow{2}{*}{Depth Map}
      & SimpleView \cite{goyal2021revisiting}        & 72.3 & 68.8 & 57.2 & 63.3  & 49.2 & 79.7 & 62.5 & 66.5 & 64.8 & 85.8 \\
      & LidarGait \cite{shen2023lidargait}   & 91.8 & 88.6 & 74.6 & 89.0 & 67.5 & 94.5 & 80.9 & 90.4 & 86.8 & 96.1 \\
    \midrule
    \multirow{3}{*}{Point Cloud}
      & HMRNet \cite{han2024gait}     & 92.7 & 92.3 & 79.6 & 90.3 & 83.1 & 95.2 & 86.2 & 90.4 & 90.2 & 97.5  \\
      & PointNet++ \cite{qi2017pointnet++}   & 82.5 & 78.7 & 58.7 & 76.1 & 74.0 & 85.4 & 75.8 & 74.8 & 77.1 & 94.1 \\
      & LidarGait++ \cite{shen2025lidargait++}   & 94.2 & 93.9 & \textbf{79.7} & 92.4 & 91.5 & \textbf{96.6} & 91.9 & \textbf{92.2} & 92.7 & \textbf{98.2} \\
    \midrule
    \multirow{3}{*}{Silhouette+Skeleton}
      & BiFusion  \cite{peng2024learning}       & 69.8 & 62.3 & 45.4 & 60.9 & 54.3 & 63.5 & 77.8 & 33.7 & 62.1 & 83.4 \\
      & SkeletonGait++ \cite{fan2024skeletongait}    & 85.1 & 82.9 & 46.6 & 81.9 & 80.8 & 82.5 & 86.2 & 47.5 & 81.3 & 95.5 \\
      & PSGait  \cite{xu2025psgait}         & 87.9 & 85.2 & 53.9 & 84.5 & 86.3 & 85.3 & 91.4 & 47.5 & 84.6 & 94.8 \\
    \midrule


    \multirow{2}{*}{Silhouette+Parsing}
      & XGait   \cite{zheng2024takes}         &  90.2 & 86.6 & 45.6 & 87.8 & 82.3 & 85.6 &  92.2 & 48.5 & 86.2 & 94.9 \\
      & MultiGait++  \cite{jin2025exploring}     & 93.1 & 88.4 & 38.7 & 89.2 & 88.1 & 88.4 & 94.6 & 48.1 & 87.9 & 95.3 \\
    \midrule

    
    \multirow{2}{*}{Sketch+Parsing}
      & MultiGait++       & 96.6 & 92.9 & 53.7 & 92.7 & 92.9 & 92.0 & 98.4 & 53.1 & 91.9 & 97.1 \\
      & SketchGait (Ours)        & 97.1 & 94.2 & 60.3 & 93.5 & 93.9 & 93.7 & 98.8 & 55.1 & 92.9 & 97.3 \\
      \midrule
     \multirow{1}{*}{Sket.+Silh.+Pars.}
      & SketchGait++ (Ours)        & \textbf{97.5} & \textbf{94.4} & 60.1 & \textbf{93.6} & \textbf{94.1} & 93.9 & \textbf{98.9} & 54.0 & \textbf{93.1} & 97.4 \\

    \bottomrule
  \end{tabular}
\end{table*}
\subsection{Implementation Details}
All models are trained using SGD with momentum 0.9 and weight decay $5\times10^{-4}$. The initial learning rate is set to 0.1 and is decayed by a factor of 0.1 at predefined milestones.
On \textbf{SUSTech1K}, we train for 50K iterations with a batch size of $8\times8\times10$, denoting 8 identities, 8 types per identity, and 10 sequences per type. The learning rate decays at 20K, 30K, and 40K iterations. On \textbf{CCPG}, we train for 120K iterations with a batch size of $32\times2\times30$, with learning rate decay at 40K, 80K, and 100K iterations.
\subsection{Effectiveness of Sketch as a Gait Modality}
\textbf{Choice of edge-based detectors.} To comprehensively evaluate different edge extraction methods, we examine four representative detectors on SUSTech1K, including two traditional methods, Canny \cite{canny2009computational} and Sobel \cite{sobel19683x3}, as well as two recent state-of-the-art methods, PiDiNet \cite{su2021pixel} and TEED \cite{yang2024deep}.

\textbf{Overall Comparison Across Visual Representations on SUSTech1K.}
As shown in Table~\ref{tab:modalities_results}, we have three main observations.
(1) In the silhouette setting, the silhouette generated from \textbf{M2FP} \cite{yang2024deep} performs better than the official released silhouette modality  under the same backbone. This suggests that more accurate foreground boundaries can further improve silhouette-based gait recognition.
(2) For the parsing modality, previously adopted parsing models in gait recognition, including GaitParsing and CDGNet, perform notably worse than the stronger human parsing model M2FP. This result indicates that parsing performance is highly dependent on segmentation quality and part granularity.
(3) The proposed Sobel-based and TEED-based \textbf{Sketch} representations both outperform silhouette-based representations, achieving OA@R-1 scores of 87.1 (+\textbf{6.2\%}) and 89.6 (+\textbf{8.7\%}), respectively. Among them, TEED-based \textbf{Sketch} performs best, verifying the effectiveness of dense, \textit{label-free} structural cues for gait representation learning.

\textbf{Effect of Different Sketch Construction Methods.}
To examine whether the effectiveness of Sketch depends on a specific edge detector, we compare several Sketch construction methods, including both traditional operators and deep learning-based detectors. As shown in Table~\ref{tab:modalities_results}, different construction methods lead to different performance levels, while all of them support Sketch as a viable gait modality.

In our experiments, Sobel and TEED achieve the most stable performance, suggesting that they preserve richer and more reliable structural edges for gait recognition. 
By contrast, Canny and PiDiNet sometimes show lower stability or fail to capture certain fine edge details, suggesting that their edge maps may be less reliable for gait representation.
These observations indicate that the effectiveness of Sketch depends on the generation quality of the adopted construction method.

\textbf{Analysis of Parsing-derived Silhouette and Sketch.}
For a fair comparison, we derive both \textit{Silhouette} and \textit{Sketch} from the outputs of three parsing models. Removing semantic labels while preserving dense structural boundaries produces  \textit{Sketch}, whereas removing internal structures and keeping only a single foreground label produces \textit{Silhouette}.

Table~\ref{tab:modalities_results} shows a clear trend for both GaitBase and DeepGaitV2: parsing-derived \textit{Sketch} outperforms parsing-derived \textit{Silhouette} in most cases, suggesting that dense, \textit{label-free} structural information is more effective than a low-entropy contour representation. However, their advantages vary across conditions. \textit{Sketch} achieves higher Rank-1 accuracy on \textbf{NM} and overall settings, while \textit{Silhouette} performs better on \textbf{CL}; for \textbf{NT}, neither shows a stable advantage. This aligns with their characteristics: \textit{Silhouette} favors global shape consistency and is more stable under \textbf{CL}, whereas \textit{Sketch} preserves richer internal structures and finer boundaries, leading to stronger overall performance. A similar pattern is observed for the M2FP-derived representations, further supporting \textit{Sketch} as a dense, \textit{label-free} structural modality.

\begin{table}[t]
  \centering
  \caption{Performance comparison on CCPG.}
  \label{tab:ccpg_comparsion}
  \small
  \setlength{\tabcolsep}{4pt}
  \renewcommand{\arraystretch}{1.0}
  \begin{tabular}{llccccc}
    \toprule
    \textbf{Input} & \textbf{Method} &
    \textbf{CL} & \textbf{UP} & \textbf{DN} & \textbf{BG} & \textbf{Mean} \\
    \midrule

    \multirow{4}{*}{Silhouette}
      & GaitSet      & 60.2  & 65.2 & 65.1 & 68.5 & 64.8 \\
      & GaitPart     & 64.3 & 67.8   & 68.6 & 71.7 & 68.1 \\
      & GaitBase     & 71.6 & 75.0 & 76.8 & 78.6 & 75.5 \\
      & DeepGaitV2   & 78.6 & 84.8 & 80.7 & 89.2 & 83.3 \\
    \midrule
 \multirow{2}{*}{Skeleton}
      & GaitGraph2    &  5.0 &  5.3 & 5.8 & 6.2  & 5.6 \\
      & SkeletonGait   &  40.4   &  48.5 & 53.0 & 61.7 & 50.9 \\
    \midrule
  
    \multirow{5}{*}{Sketch}
      & & \multicolumn{5}{c}{\cellcolor{m2fpblue}\textit{M2FP-to-Edge}} \\
      & DeepGaitV2
      & \cellcolor{m2fpblue}63.2
      & \cellcolor{m2fpblue}71.0
      & \cellcolor{m2fpblue}76.6
      & \cellcolor{m2fpblue}84.0
      & \cellcolor{m2fpblue}73.7 \\
      \addlinespace[2pt]
      & & \multicolumn{5}{c}{\cellcolor{teedgreen}\textit{TEED}} \\
      & DeepGaitV2
      & \cellcolor{teedgreen}61.0
      & \cellcolor{teedgreen}67.4
      & \cellcolor{teedgreen}75.7
      & \cellcolor{teedgreen}83.1
      & \cellcolor{teedgreen}71.2 \\
      \addlinespace[2pt]

    \midrule
   
    \multirow{1}{*}{Parsing}
      & DeepGaitV2   &  68.8 &  73.6 & 80.1 & 87.8 & 77.6 \\
      
    \midrule

    \multirow{3}{*}{Silh.+Skel.}
      & BiFusion       & 62.6 & 67.6 & 66.3 & 66.0 & 65.6 \\
      & SkeletonGait++ & 79.1 & 83.9 & 81.7 & 89.9 & 83.7 \\
      & PSGait & \textbf{82.5} & 85.3 & 88.6 & 89.1 & 86.4 \\
    \midrule

    \multirow{2}{*}{Silh.+Pars.}
      & XGait        & 73.9  & 75.5   &  79.8   & 81.2   & 77.6   \\
      & MultiGait++  &  79.4  & 83.2   & 83.7   &  90.9   & 84.3   \\
    \midrule


    \multirow{5}{*}{Sket.+Pars.}
      & & \multicolumn{5}{c}{\cellcolor{m2fpblue}\textit{M2FP-to-Edge}} \\
      & SketchGait
      & \cellcolor{m2fpblue}74.8
      & \cellcolor{m2fpblue}80.9
      & \cellcolor{m2fpblue}83.8
      & \cellcolor{m2fpblue}90.5
      & \cellcolor{m2fpblue}82.5 \\
      \addlinespace[2pt]
      & & \multicolumn{5}{c}{\cellcolor{teedgreen}\textit{TEED}} \\
      & MultiGait++
      & \cellcolor{teedgreen}69.9
      & \cellcolor{teedgreen}77.3
      & \cellcolor{teedgreen}80.2
      & \cellcolor{teedgreen}91.0
      & \cellcolor{teedgreen}79.6 \\
      & SketchGait
      & \cellcolor{teedgreen} 75.3
      & \cellcolor{teedgreen}81.1
      & \cellcolor{teedgreen}84.5
      & \cellcolor{teedgreen}92.8
      & \cellcolor{teedgreen}83.4 \\
\midrule
       \multirow{2}{*}{Sket.+Silh.+Pars.}
      
      & & \multicolumn{5}{c}{\cellcolor{teedgreen}\textit{TEED}} \\
      & SketchGait++
      & \cellcolor{teedgreen} 82.2
      & \cellcolor{teedgreen}\textbf{87.0}
      & \cellcolor{teedgreen}\textbf{86.8}
      & \cellcolor{teedgreen}\textbf{94.5}
      & \cellcolor{teedgreen}\textbf{87.6} \\

    \bottomrule
  \end{tabular}
\end{table}

\subsection{Comparison with State-of-the-Art Methods.}
As shown in Table~\ref{tab:sustech1k_comparsion}, multi-modal settings with \textit{Sketch} consistently outperform other modality combinations. In particular, \textit{Sketch+Parsing} yields the best performance among all bi-modal settings. MultiGait++ and SketchGait achieve overall Rank-1 accuracies of \textbf{91.9}\% and \textbf{92.9}\%, respectively, both outperforming the silhouette+parsing setting. By further introducing the silhouette modality, SketchGait++ reaches \textbf{93.1}\%.

On CCPG, SketchGait achieves a mean Rank-1 accuracy of only \textbf{83.4}\%, reflecting limitations of the Sketch modality. However, by introducing the silhouette modality, SketchGait++ shows a substantial improvement, reaching \textbf{87.6}\% and outperforming other multi-modal methods, as shown in Table~\ref{tab:ccpg_comparsion}.

\subsection{Ablation Study of SketchGait}
We conduct ablation experiments on SketchGait from two aspects: (1) whether multi-branch modeling is necessary for jointly exploiting Sketch and Parsing, and (2) how different fusion stages and fusion operators influence recognition performance. The results are reported in Table~\ref{tab:ablation_fusion_settings}.

\textbf{Necessity of Multi-branch Modeling.}
We first evaluate whether Sketch and Parsing should be modeled with independent branches. Compared with the single-modality Sketch and Parsing baselines, the dual-branch non-fusion baseline improves the overall performance (OA@R-1) by \textbf{3.1}\% and \textbf{5.2}\%, respectively. Moreover, it outperforms the \textit{Concat-only} single-branch variant by \textbf{4.0}\%.

These results show that jointly using Sketch and Parsing is beneficial, and that preserving two independent modality-specific branches is more effective than directly concatenating the two modalities into a single branch. This supports the necessity of multi-branch modeling in SketchGait.

\textbf{Effect of Fusion Stage and Fusion Operator.}
We further study how the fusion stage and fusion operator influence performance. Under the \textbf{Add} operator, early-stage fusion (Stage-1) performs better than mid-stage fusion (Stage-3), indicating that Sketch and Parsing interact more effectively at shallow stages. In contrast, the fused-only early-stage variant performs much worse, suggesting that early interaction is not sufficient and should be combined with independent modality-specific branches.

We also compare different fusion operators under the early-stage setting. \textbf{Add} achieves performance comparable to \textbf{Concat} and \textbf{Attention}, showing that SketchGait does not rely on a complex fusion module to obtain strong results.

\begin{table}[t]
  \centering
  \caption{Ablation study of different fusion configurations on SUSTech1K. The compared settings include non-fusion, concat-only, mid-stage fusion (Stage-3), and two early-stage fusion variants (fused-only and tri-branch, both at Stage-1).}
  \label{tab:ablation_fusion_settings}
  \small
  \setlength{\tabcolsep}{5pt}
  \renewcommand{\arraystretch}{1.08}
  \begin{tabular}{llcccc}
    \toprule
    \textbf{Modality} & \textbf{Setting} & \textbf{NM} & \textbf{CL} & \textbf{NT} & \textbf{OA@R-1} \\
    \midrule
    Parsing
      & DeepGaitV2 & 93.1 & 40.9 & 47.5 & 87.5 \\
    \midrule

    Sketch
      & DeepGaitV2 & 94.5 & 56.4 & 51.5 & 89.6 \\
    \midrule

    \multirow{17}{*}{Sketch+Parsing}
      & \multicolumn{5}{l}{\textit{Non-fusion (dual-branch)}} \\
      \cmidrule(lr){2-2}
      & Baseline             & 97.0 & \textbf{60.5} & 53.7 & 92.7 \\

      & \multicolumn{5}{l}{\textit{Concat-only (single branch)}} \\
      \cmidrule(lr){2-2}
      & DeepGaitV2           & 93.5 & 42.4 & 51.3 & 88.7 \\
      \addlinespace[2pt]

      & \multicolumn{5}{l}{\textit{Mid-stage fusion (tri-branch, Stage-3)}} \\
      \cmidrule(lr){2-2}
      & Attention            & 97.0 & 58.2 & 54.2 & 92.5 \\
      & Concat               & 97.2 & 60.0 & 54.7 & 92.8 \\
      & Add                  & \textbf{97.3} & 59.1 & 54.4 & 92.8 \\
      \addlinespace[2pt]

      & \multicolumn{5}{l}{\textit{Early-stage fusion (fused-only, Stage-1)}} \\
      \cmidrule(lr){2-2}
      & Add                  & 96.0 & 51.1 & 52.7 & 91.3 \\
      \addlinespace[2pt]

      & \multicolumn{5}{l}{\textit{Early-stage fusion (tri-branch, Stage-1)}} \\
      \cmidrule(lr){2-2}
      & Attention            & 97.1 & 60.3 & 54.7 & 92.9 \\
      & Concat               & 97.2 & 60.0 & 54.6 & 92.9 \\
      & \textbf{Add (ours)}  & 97.1 & 60.3 & \textbf{55.1} & \textbf{92.9} \\
    \bottomrule
  \end{tabular}
\end{table}

\subsection{Complementarity Between Sketch and Parsing}
\textbf{Structural Edge Complementarity.}
As shown in Table~\ref{tab:sustech1k_comparsion} and Table~\ref{tab:ccpg_comparsion}, combining \textit{Sketch} and \textit{Parsing} consistently outperforms using either modality alone. On SUSTech1K, SketchGait achieves \textbf{92.9}\%, surpassing the parsing modality (\textbf{87.5}\%) and the sketch modality (\textbf{89.6}\%) by \textbf{5.4}\% and \textbf{3.3}\%, respectively. On CCPG, SketchGait reaches \textbf{83.4}\%, outperforming the parsing modality (\textbf{77.6}\%) and the sketch modality (\textbf{71.2}\%) by \textbf{5.8}\% and \textbf{12.2}\%, respectively. These results indicate that \textit{Sketch} and \textit{Parsing} provide complementary cues for gait recognition.

A possible explanation is that the two modalities capture different structural information. \textit{Sketch} preserves rich high-frequency boundaries and internal contours, while \textit{Parsing} mainly captures semantic part structure. Under self-occlusion, \textit{Parsing} may miss fine details due to coarse-grained body partitioning, whereas \textit{Sketch} preserves clearer local boundaries. Their complementarity makes the combination effective for gait recognition.

\textbf{Parsing Reduces Texture-related Interference on CCPG.   }
As shown in Table~\ref{tab:ccpg_comparsion}, TEED-based  \textit{Sketch} achieves only \textbf{71.2}\% mean Rank-1 on CCPG, substantially lower than its performance on SUSTech1K (Table~\ref{tab:modalities_results}). The degradation is particularly evident under clothing-related settings, especially \textbf{CL} (61.0\%) and \textbf{UP} (67.4\%). In contrast, M2FP-to-Edge \textit{Sketch} performs better, suggesting that parsing-derived edges are less affected by complex clothing logos and texture patterns. Taken together, these results indicate that TEED-based \textit{Sketch} is more susceptible to texture-induced noises on CCPG, where rich clothing appearance introduces high-frequency edges that are not reliable for gait recognition.

Despite this limitation, TEED-based \textit{Sketch} with M2FP \textit{Parsing} still outperforms M2FP-to-Edge with \textit{Parsing}. This suggests that the complementarity between \textit{Sketch} and \textit{Parsing} is not limited to structural information. Instead, \textit{Parsing} provides more effective complementary cues for the finer-grained TEED-based \textit{Sketch}, which may help mitigate texture-related interferences in the \textit{Sketch} modality.

\begin{table}[t]
\centering
\caption{Cross-dataset generalization results. Left: trained on CCPG and tested on SUSTech1K. Right: trained on SUSTech1K and tested on CCPG.}
\label{tab:cross_dataset_generalization}
\setlength{\tabcolsep}{2.8pt}
\footnotesize
\resizebox{\columnwidth}{!}{
\begin{tabular}{ll|cccc|cccc}
\toprule
\multirow{2}{*}{Input} & \multirow{2}{*}{Method} 
& \multicolumn{4}{c|}{CCPG $\rightarrow$ SUSTech1K} 
& \multicolumn{4}{c}{SUSTech1K $\rightarrow$ CCPG} \\
\cmidrule(lr){3-6} \cmidrule(l){7-10}
& & NM & CL & NT & OA@R-1 & CL & UP & DN & BG \\
\midrule
Silhouette   & DeepGaitV2 & 27.2   & 17.6   &14.7   & 23.0   & \textbf{13.0 }  & 17.9  & 26.8   &  39.4   \\
Sketch  & DeepGaitV2 & 37.5 & 20.7 & 20.3 & 35.0 & 8.1  & 14.6 & 22.5 & 48.9 \\
Parsing & DeepGaitV2 & 51.8 & 24.3 & 22.6 & 46.3 & 11.0 & 17.7 & 32.8 & 60.0 \\
\midrule

Sils.+Pars. & MultiGait++ & 44.3 & 18.3 & 18.5 & 36.7 & 11.3& 17.4 & 32.8 & 54.3 \\
Sket.+Pars.  & MultiGait++ & 57.0 & 25.6 & 24.1 & 52.0 & 8.6 & 17.1 & 28.1 &60.4 \\
 & \textbf{SketchGait}  & \textbf{61.0} & \textbf{33.1 }& \textbf{27.8} & \textbf{55.9} & 10.3  & 19.7& 32.8 & 67.9 \\
 \midrule
 Sket.+Pars.+Sils.  & SketchGait++ &  57.9 & 30.0 & 24.6 & 51.5 &12.3 & \textbf{22.1 } & \textbf{34.2} &\textbf{70.3} \\

\bottomrule
\end{tabular}
}
\end{table}

\section{Cross-domain Comparison}
Table~\ref{tab:cross_dataset_generalization} reports the cross-dataset results. Three observations are worth noting. First, among the three visual modalities, Parsing achieves the strongest overall generalization performance on both datasets. Second, in the CCPG $\rightarrow$ SUSTech1K setting, SketchGait obtains the best results across all metrics, showing clear advantages over the competing methods. In the SUSTech1K $\rightarrow$ CCPG setting,  SketchGait++ performs better.
\begin{figure}[h]

  \centering
  \includegraphics[width=\linewidth]{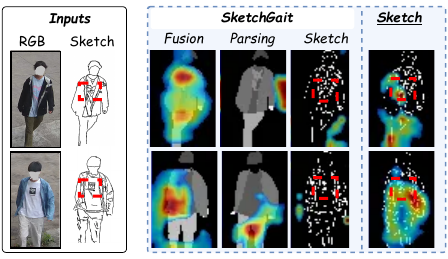}
  
\caption{Example frames from  CCPG. \underline{\textbf{\textit{Sketch}}} shows the heatmap of DeepGaitV2 when \textit{Sketch} is used as the only input modality.
}
\label{fig:heatmap}
\end{figure}
\section{Visualization}
As shown in Fig.~\ref{fig:heatmap}, the \textbf{\textit{Sketch-based}} method often highlights body textures and clothing logos, which may lead to poor generalization. When combined with Parsing in SketchGait, the \textit{Sketch} branch shows a clear reduction in reliance on complex upper-body textures. We also observe that the \textit{Fusion} branch activates broadly across the entire body, indicating clear complementarity between the two modalities and supporting the effectiveness of early-stage fusion.

\section{Limitations and Future Work}

\textbf{Limitations.}
Although the combination of \textit{Sketch} and \textit{Parsing} achieves strong performance, \textit{Parsing} remains more reliable than \textit{Sketch} when the two modalities are evaluated separately. We attribute this mainly to the following reason.

\textbf{Over-extraction of clothing textures.}  
Current edge-based detectors are not designed for gait recognition and may extract texture edges irrelevant to identity-related motion cues. In contrast, \textit{Parsing} represents the body via semantic part regions, making it less sensitive to clothing logos and dense textures, and thus a more stable and reliable visual modality.

\textbf{Future Work.}
First, since \textit{Sketch} has shown clear potential for gait recognition, future work can focus on reducing texture interference. This may be explored from three aspects: \textbf{(1)} developing gait-oriented edge detectors to suppress irrelevant texture edges; \textbf{(2)} designing more effective pre-processing methods to reduce clothing logos and dense texture patterns before edge extraction; and \textbf{(3)} introducing denoising modules or better multi-modal designs to alleviate texture-related interference during feature learning.

Second, effective fusion of different visual modalities remains important. Current multi-modal gait recognition mainly focuses on skeleton, silhouette, and parsing. With the introduction of \textit{Sketch}, a key question is how to better integrate it with these modalities in a unified framework. We believe that more effective \textit{Sketch}-based multi-modal modeling is a promising direction for gait recognition.

\section{Conclusion}

In this paper, we revisit gait visual representations from two perspectives: structural edge density and semantic richness. This view provides a unified understanding of existing gait modalities and reveals an underexplored direction: dense structural cues without explicit semantic labels. Based on this, we introduce \textbf{Sketch} as a new gait modality. As a dense, \textit{label-free} representation extracted from RGB images, Sketch preserves rich structural cues and complements Parsing for gait recognition. To exploit this, we propose \textbf{SketchGait}, a multi-modal framework with two modality-specific branches and a lightweight early-stage fusion branch. By introducing the silhouette modality, we extend it to \textbf{SketchGait++} for better performance. Experiments on SUSTech1K and CCPG demonstrate the effectiveness of both Sketch and the framework. The results show that Sketch is a strong structural representation, and combining it with Parsing consistently improves recognition. Results on CCPG also suggest that edge-based Sketch may suffer from texture-related interference under clothing changes, pointing to future work on reducing texture noise and improving multi-modal fusion.

\bibliographystyle{ACM-Reference-Format}
\bibliography{sketchgait}

\end{document}